\documentclass[12pt, a4paper]{article}
\usepackage{amsmath}
\usepackage{graphicx}
\usepackage{geometry}
\usepackage{setspace}
\usepackage{tocloft} 
\usepackage{lipsum}
\usepackage{mathptmx} 
\usepackage[utf8x]{inputenc}
\usepackage{hyperref}
\usepackage[notocbib]{apacite}
\usepackage{colortbl} 
\usepackage{xcolor}   
\usepackage{makecell}
\usepackage{multicol}
\usepackage{multirow}
\usepackage{booktabs} 
\usepackage{array}    
\usepackage{tabularx} 
\usepackage{caption}  
\usepackage{enumitem}
\usepackage{float}
\usepackage{indentfirst}
\usepackage{amsmath}
\usepackage{titlesec}

\titleformat*{\subsubsection}{\large\bfseries}

\usepackage{graphicx}
\graphicspath{ {./images/} }

\renewcommand{\arraystretch}{1.5}
\author{}
\date{}

\usepackage{times}
\begin{document}
\pagenumbering{roman}
\addcontentsline{toc}{section}{\protect\numberline{}Title page}
\begin{center}
\large B.Comp. Dissertation
\text{ }\\
\text{ }\\
\Large \textbf{Enhancing Sports Strategy with Video Analytics and Data Mining} \\
\large Assessing the effectiveness of Multimodal LLMs in tennis video analysis
\vfill
\large By\\
\large Teo Wei Jun Charlton\\
\vfill
\large Department of Computer Science \\
\large School of Computing\\
\large National University of Singapore\\
\large 2024/2025
\end{center}
\text{ }\\
\large Project No: H403040\\
\large Advisor: Jiang Kan
\vfill

\newpage

\addcontentsline{toc}{section}{\protect\numberline{}Abstract}
\begin{center}
    \large Abstract
\end{center}
The use of Large Language Models (LLMs) in recent years has also given rise to the development of Multimodal LLMs (MLLMs). These new MLLMs allow us to process images, videos and even audio alongside textual inputs. In this project, we aim to assess the effectiveness of MLLMs in analysing sports videos, focussing mainly on tennis videos. Despite research done on tennis analysis, there remains a gap in models that are able to understand and identify the sequence of events in a tennis rally, which would be useful in other fields of sports analytics. As such, we will mainly assess the MLLMs on their ability to fill this gap - to classify tennis actions, as well as their ability to identify these actions in a sequence of tennis actions in a rally. We further looked into ways we can improve the MLLMs' performance, including different training methods and even using them together with other traditional models.
\\
\\
Subject Descriptors:\\
\\
\hspace*{15mm} I.2.7 \space \space \space \space \space Natural Language Processing\\
\hspace*{15mm} I.2.10 \space \space \space Vision and Scene Understanding\\
\hspace*{15mm} I.4 \space \space \space \space \space \space \space \space Image Processing and Computer Vision\\
\\
Keywords:\\
\\
\hspace*{15mm} Sports Analytics, sequence identification, multimodal large language 
\hspace*{15mm} models, pose detection, object detection
\newpage

\addcontentsline{toc}{section}{\protect\numberline{}Acknowledgments}
\begin{center}
    \large Acknowledgments
\end{center}
I would like to thank my supervisor, Prof Jiang Kan, research assistant, Liu Zhaoyu, UROP undergraduate, Ervin Yeoh, and my assessor, Prof Eldon Chung, for their endless support and continued guidance throughout this final year project. Without their inputs and help, this project would not have been possible. 
\newpage

\pagenumbering{gobble}
\tableofcontents
\newpage
\renewcommand{\arraystretch}{1.5}

\pagenumbering{arabic}

\section{Introduction}
\subsection{Multimodal Large Language Models (MLLMs)}

In recent years, there has been heavy competition in the field of LLMs, from ChatGPT to DeepSeek. On top of interpreting language, these models have also begun developing to take in other modes of information such as videos, images and even audio, developing into MLLMs. A lot of debate has been sparked on which of these MLLMs are the best, and in which different situations do they thrive. However, since a lot of these MLLMs have been pre-trained on large amounts of data, they are usually generic and are not as effective in topic-specific prompts. As such, we are interested to see how MLLMs can be used in a more specific, niche purpose. 

\subsection{Sports analytics}

Sports analytics has become increasingly prevalent in the sporting world today. Many sports teams and players in varying sports have data analytics teams to analyse their opponents' historical data to gain an advantage and find weaknesses in their game play. Sports analytics involves various types of model and methods, and these are being continually developed to this day. With the development of MLLMs, this opens many new opportunity and avenues for us to leverage this technology in the field of sports analytics.
\newpage
\section{Literature Review}
\subsection{Multimodal Large Language Models (MLLMs)}
The field of MLLMs is still ever-evolving as new models continue to outperform one another. The main challenge at hand is how to effectively integrate different modes of data into the same representation subspace, also known as modality alignment. The development of Contrastive Language-Image Pre-Training (CLIP) has allowed for major developments in this area, showing impressive zero-shot learning, which has resulted in many MLLMs making use of its text and image encoders in their architecture. \cite{li2024surveying}

Despite strides in model developments, the current metrics of assessing the accuracy of MLLMs are still limited, focussing largely on Multiple-Choice Video Question Answering (MC-VQA). While there are some benchmarks which provide more complexity in evaluations, they are still often heavily simplified. For instance, ActivityNet \cite{yu2019activitynet} is an open-ended VQA benchmark, no longer giving options. However, the maximum length of a reply is capped at 5 words, with all answers averaging below 2 words. Table 1 below categorises some of the more common benchmarks used to evaluate recent MLLMs and highlights the limited scope they cover. These benchmarks are cross-referenced across various open-source MLLM papers, namely VideoLLaMA2 \cite{cheng2024videollama}, Oryx \cite{liu2024oryx}, VideoLLaVA \cite{lin2023video} and Aria \cite{li2024aria}.
\begin{table}[h!]
\begin{center}
\begin{tabular}{ c|c } 
 \hline
 MC-VQA & \makecell{MVBench \cite{li2024mvbench}, VideoMME \cite{fu2024video} \\ PerceptionTest \cite{patraucean2024perception} \\ Egoschema \cite{mangalam2023egoschema}, \\ 
 LongVideoBench \cite{wu2024longvideobench}}\\  
 \hline
 Video captioning & MSVD-QA \cite{chen2011collecting} \\
 \hline
 Open-ended VQA & ActivityNet \cite{yu2019activitynet}  \\
 \hline
\end{tabular}
\caption{Common benchmarks for MLLMs categorised by modality}
\end{center}
\end{table}

\par
Such sentiments are shared by other papers as well, describing current benchmarks as overly simple and uninformative, \cite{zhu2024lime} further highlighting the need for more rigorous tests and comparisons of MLLMs. 

\par
Due to their recency, there is also a lack of papers detailing attempts to fine-tune these models for their own use cases, although there are some which fine-tune on more general datasets but do not display very promising results \cite{zhai2024investigating}. These MLLM models are often used in a zero shot settings and are still being continually developed, with some, such as Oryx, still working on releasing documentation on how to fine-tune their model. 
\par
As such, it is clear that MLLMs are still early in development, and have a lot more to be researched about. On first glance, MLLMs appear to be a good fit to solve the gap in tennis analytics mentioned above. However, their current scores on relatively simplified metrics, compared to sequence analysis, begs the question if MLLMs will perform well on this task, and if it is possible to adapt them to be better suited for this task.
\subsection{Sports analytics}

Recent works in sports analytics have explored a wide range of challenges, including strategy modeling using probabilistic reasoning and deep learning \cite{dong2023sports,liu2023insight,liu2023recognizing,liu2024pcsp,liu2024exploring,liu2024strategy,hundal2024soccer}, injury prediction \cite{liu2023sports}, fine-grained event detection in sports videos \cite{liu2025f}, and specific tasks such as court detection and ball tracking in broadcast footage \cite{jiang2020deep,jiang2023court,jiang2024tracking}. These efforts demonstrate the growing integration of AI, data mining, and formal methods in advancing sports performance analysis and tactical understanding \cite{liu2025analyzing}.

\subsection{LLMs in sports}

LLMs have found multiple purposes in the world of sports, including automatic commentary generation and training motor skills. While LLMs have generally shown promise, there are still multiple issues to be seen when using them in these cases. 

In the training of motor skills, it was found that athletes coached under LLMs showed steady improvements, indicating LLMs' abilities in sports-related tasks. Nonetheless, the experiment also found that the LLMs hallucinate often, providing feedback that was completely irrelevant to badminton and giving fake information, resulting in worse performances compared to university coaches. \cite{qiu2024impact} 

In the paper on commentary generation, while the LLM did show promise, it was still far from that of a human, citing LLMs' problems with dealing with unseen events and sequence of events, which is very common in sports. \cite{cook2024llm} 

Another paper on predicting tactical decisions in football found that while LLMs gave mostly plausible answers, they still ended up being largely wrong, further highlighting that there is still work to be done to develop LLMs. \cite{caron2023tacticalgpt}

In general, LLMs have shown great promise in aiding the field of sports, making processes more efficient but are still far from replacing humans within the field, especially in terms of details and accuracy. 

\subsection{MLLMs in sports}

Given the relative recency of MLLMs, little literature is out there, especially for sports. The main first step into MLLM research for sports is SPORTU-video. It is the first dataset designed to assess MLLM's understanding of sports, comprising 1701 videos across 7 different sports. It is also important to note that this dataset, similar to what was mentioned abve, is largely multiple choice questions, with only 1075 out of 12048 questions being open-ended. These questions also vary in difficulty, from basic recognition to understanding faults and rules in the sport.

On this new dataset, the accuracy of the models varied greatly. Closed-source models varied in accuracy from 58.19 to 70.18, while open-source models varied in accuracy from 46.39 to 70.94, indicating some room for improvement.

Overall, the study reported that most of the MLLMs did well in textual analysis and reasoning but did poorly when it came to connecting images with sporting rules and analysing scenarios. 

Other papers have also cited rather poor performances, albeit for off-the-shelf MLLMs in sports. In one study, MLLMs were tasked to identify the key person in an NCAA basketball images, i.e. the person holding the ball. The two off-the-shelf MLLMs performed poorly, obtaining mAP scores of 25.95 and 32.39, which were far off state-of-the-art models. \cite{madan2024mip}

In the next section, we will similarly assess MLLMs' abilities to understand sports, mainly focussing on tennis. 
\newpage
\section{Experiments}

\subsection{Dataset}
The dataset we will be using is FineTennis \cite{liu2025f}, a new tennis video dataset detailing the sequence of events occurring in numerous rallies across several years and across the famous tournaments in tennis. \\

Each event in a rally is classified by 5 sub-classes, which we will denote as \begin{math}
    e_1, e_2, e_3, e_4
\end{math} and \begin{math}
    e_5
\end{math} as explained below: 
\begin{itemize}
\item \begin{math}
    e_1
\end{math} - which player hit the ball: \textit{near} or \textit{far}
\item \begin{math}
    e_2
\end{math} - how the ball was hit: \textit{forehand} or \textit{backhand}
\item \begin{math}
    e_3
\end{math} - shot type: \textit{serve}, \textit{return} or \textit{stroke}
\item \begin{math}
    e_4
\end{math} - shot direction: \textit{T}, \textit{body (B)} or \textit{wide (W)} for serves and \textit{cross-court (CC)}, \textit{down the line (DL)}, \textit{down the middle (DM)}, \textit{inside out (IO)} or \textit{inside in (II)} for returns and strokes
\item \begin{math}
    e_5
\end{math} - shot outcome: \textit{in} or \textit{last}

\end{itemize}
\par These 5 sub-classes give rise to a total of 56 possible events, comprising 7445 training videos and 2271 test videos.
\par The dataset provides the starting and ending frame number of each rally, and details which frame each event occurs in. Each rally is tagged to a match which is tagged with a YouTube link. As such, we were required to write a script to download the matches, split into their respective rallies, onto our computing cluster. 

The photos below show some examples of the frames containing the hitting moment, i.e. the frame which the event occurs.

\begin{figure}[H]
    \begin{center}
    \includegraphics[width=0.7\linewidth]{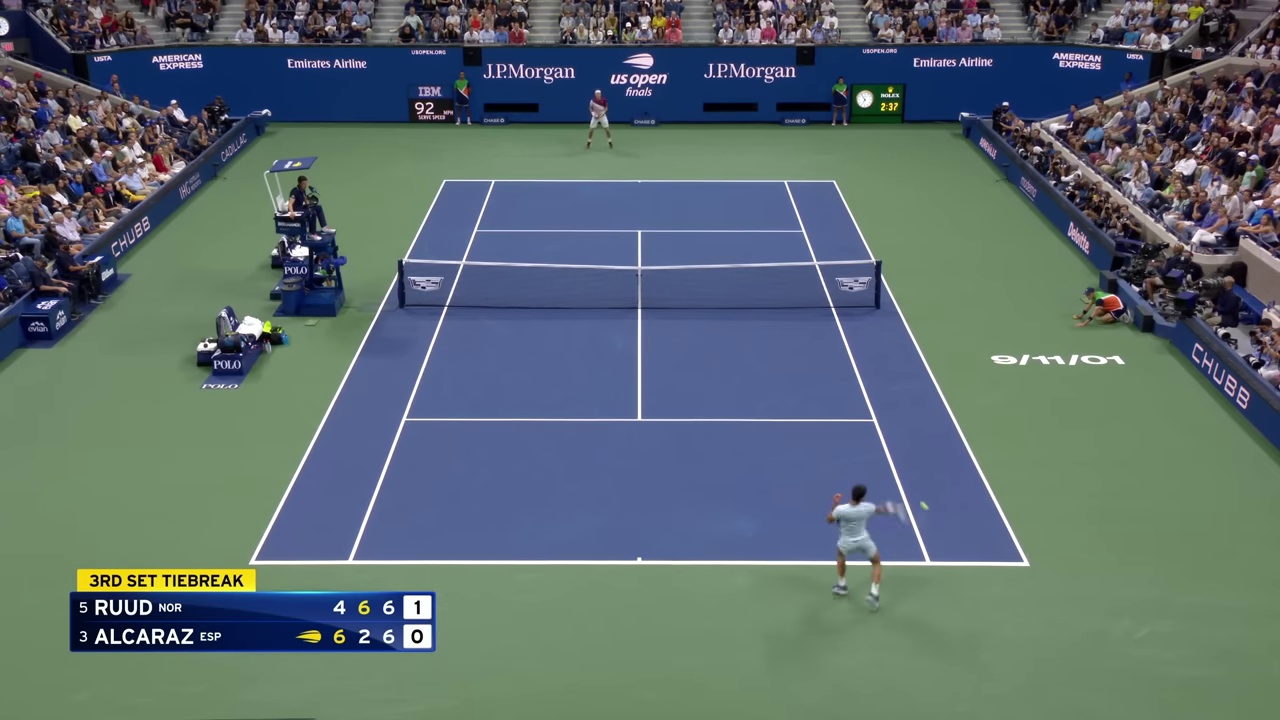}
    \caption{Sample shot in the US Open 2022 Finals}
    \label{fig:enter-label}
    \end{center}
\end{figure}

\begin{figure}[H]
    \begin{center}
    \includegraphics[width=0.7\linewidth]{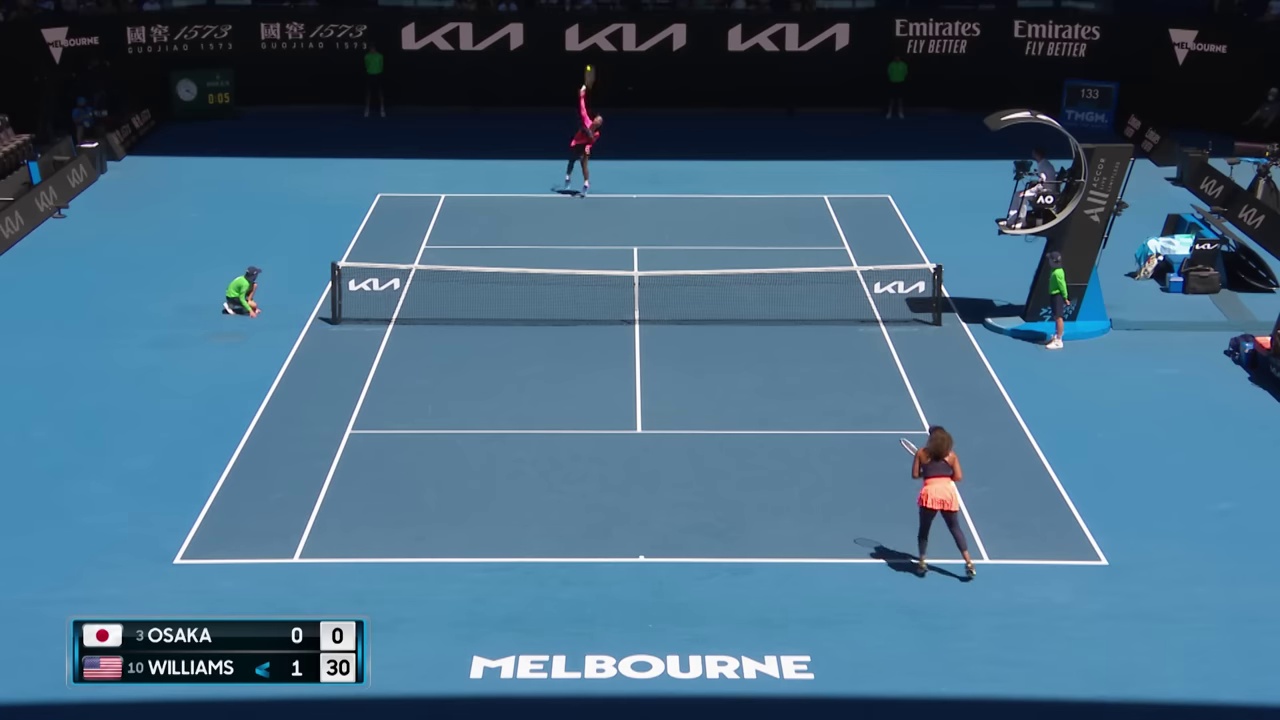}
    \caption{Sample shot in the Australian Open 2021 Semi-Finals}
    \label{fig:enter-label}
    \end{center}
\end{figure}

\par To better understand the dataset and find reasonable hyperparameters for our model in future, we did some basic analysis on the number of events for each dataset - the train set and the test set. The tables below gives more details about the training and testing dataset.

\begin{table}[h!]
\begin{center}
\begin{tabular}{ |c|c| } 
 \hline
 Mean number of events & 3.68 \\
 Lower quartile & 1 \\
 Median number of events & 3 \\
 Upper quartile & 5 \\
 Maximum number of events & 34 \\
 \hline
\end{tabular}
\caption{Statistics of training set}
\end{center}
\end{table}
\begin{table}[h!]
\begin{center}
\begin{tabular}{ |c|c| } 
 \hline
 Mean number of events & 3.90\\
 Lower quartile & 2 \\
 Median number of events & 3 \\
 Upper quartile & 5 \\
 Maximum number of events & 29 \\
 \hline
\end{tabular}
\caption{Statistics of testing set}
\end{center}
\end{table}

\newpage

\subsection{Benchmark} 

Using this dataset, we will benchmark our model against the \begin{math}F^3EST\end{math} model \cite{unpublishedkeyA}, which was evaluated on the same dataset but at different levels of granularity. For the \begin{math}F^3EST\end{math} model, it achieved an edit score of 88.4 across 38 possible event types and an edit score of 82.1 across 111 possible event types. While the granularity differs, these score still serve as a good beginning benchmark for us to evaluate our MLLM models against traditional models. The calculation of edit scores will be elaborated on further in section 3.6.

\subsection{Model Selection}
Due to the nature of MLLMs, numerous GPUs are usually required to train and fine-tune them. The only resources available to us is a shared computing cluster with limited access. Hence, we had to pick an MLLM that could be trained under these limited resources while also performing relatively well in known MLLM benchmarks, which narrowed down the possibilities. We considered some of the models mentioned above and ultimately decided on VideoLLaMA2, specifically fine-tuned with Low-Rank Adaptation (LoRA) \cite{hu2021lora}, which significantly reduces the VRAM consumption in the process. It is worth noting that using LoRA for fine-tuning might result in lower accuracy levels as compared to full fine-tuning \cite{sun2023comparative}, but it still serves as a good gauge for the ability of MLLMs to carry out the task. 

\par For the sake of consistency, we will be using the most updated VideoLLaMA2 repository as of beginning the first experiment, which would be the September 13th 2024 version. This is important as open-source MLLM repositories are being consistently upgraded and changing our own repository across experiments would result in unfair comparisons. In this version of the repository, the training script uses VideoLLaMA2-7B-Base and uses clip-vit-large-patch14-336 by OpenAI for visual encoding. 

\subsection{Single Event Classification}
Before training the model on full sequences, we decided to train the model on a smaller subtask - identifying single tennis events in videos. The FineTennis dataset is split into videos comprising only 1 singular event. To provide important context to the model, each video was split by starting 10 frames before the shot is taken and ends right before the next shot is taken. 
\par \textbf{Dataset set-up.} To feed in the labels into MLLM, the labels had to be modified to more text-based labels. In these experiments, we use the following prompts: 
\begin{itemize}
\item \textit{Prompt: } What is happening in the tennis video?
\item \textit{Answer: } The \{\begin{math}e_1\end{math}\} player hit a \{\begin{math}e_2\end{math}\} \{\begin{math}e_4\end{math}\} \{\begin{math}e_3\end{math}\} \{\begin{math}e_5\end{math}\}.
\end{itemize}
\par Since prompt engineering is a big field in LLMs, we also tried other prompt-answer pairs but found that these made little changes to the end results. Hence, we stuck to the above prompts for all experiments to ensure consistency. 
\par \textbf{Evaluation method.} Since the output might not follow a fixed format, we had to adopt another approach to determine the accuracy of the results. In our experiments, we determine that the sub-class has been predicted correctly if the correct sub-class label appears in the answer and no other possible sub-class label appears in the answer. The overall accuracy is only considered correct if all sub-classes are correctly classified. 
\par The table below details the results of our experiment, broken down by each sub-class.

\begin{table}[h!]
\begin{center}
\begin{tabular}{ |c|c| } 
 \hline
 Sub-class & Accuracy \\
 \hline
 \begin{math}e_1\end{math} & 0.96 \\
 \begin{math}e_2\end{math} & 0.73 \\
 \begin{math}e_3\end{math} & 0.83 \\
 \begin{math}e_4\end{math} & 0.72 \\
 \begin{math}e_5\end{math} & 0.96 \\
 \hline
 Overall & 0.41 \\
 \hline
\end{tabular}
\caption{Results for single event fine-tuning}
\end{center}
\end{table}
\par The results above are after the model has been fine-tuned for 8 epochs. We also experimented with 6 epochs which gave significantly worse results (0.28), and 10 epochs which results in the same accuracy (0.41). 
\par The current results are far from the benchmark score we have set, albeit the benchmark score was on a different and more complex task. 

\subsection{Rally Sequence Identification}
Next, we move on to the main task at hand. Given a video of a tennis rally, we expect the model to produce a sequence of events, detailing all events in the rally. 

\par \textbf{Dataset setup.} While the prompt in the dataset remained the same as the single event classification, the answers are now concatenations of the individual event answers. For example, for a rally video with 2 events, a \textit{near T serve in} followed by a \textit{far forehand return last}, the training reply answer is now: \textit{The near player hit a T serve in. The far player hit a forehand return last.}

\par We attempted various variations and modifications in an effort to boost the performance of the MLLM in sequence identification. These modifications do not stack. This allows them to be more fairly compared to the baseline default model. These variations include: 
\begin{enumerate}
    \item \textbf{Default (8 epochs, 10 epochs, 20 epochs, 50 epochs).} This experiment was with the default VideoLLaMA2 model with nothing changed. The dataset used is as mentioned above. To determine a suitable number of epochs for the other variations, we experimented with a range of epochs to find a balance between accuracy and time. 
    \item \textbf{Event counting.} The results of the default model indicates that VideoLLaMA2 is unable to ascertain the total number of events in a video, on top of poor single event identification. To check this, we conduct a test only on event counting, where the prompt enquires the total number of actions in the video and the answer is a single number. Given the increased simplicity of the task, we expect the model to perform better. 
    \item \textbf{Audio.} The audio was now added back to the videos in the dataset. We expect this to perform better as VideoLLaMA2 boasts advanced audio-visual integration \cite{cheng2024videollama}. Though, it is important to note that at time of testing, the repository's audio branch was not fully published. Nonetheless, the current code still seems to accept audio. The sound of the tennis racquet hitting the ball should be helpful to the model, especially in determining the number of events in a rally video.
    \item \textbf{Increased frame sampling.} One limitation we saw with VideoLLaMA2 was its sampling method. The current model samples only 8 frames per video. We increase this to 32 frames as all of the videos contain less than 32 events. With more temporal information, the model should be able to perform better. We are unable to increase the number of sampled frames too much as we are restricted by the length of the shortest video. 
    \item \textbf{Providing frame numbers.} In this variation, the answers in the training data provide the frame number for the event, i.e. each event answer changes from ``The \{\begin{math}e_1\end{math}\} player hit a \{\begin{math}e_2\end{math}\} \{\begin{math}e_4\end{math}\} \{\begin{math}e_3\end{math}\} \{\begin{math}e_5\end{math}\}." to ``The \{\begin{math}e_1\end{math}\} player hit a \{\begin{math}e_2\end{math}\} \{\begin{math}e_4\end{math}\} \{\begin{math}e_3\end{math}\} \{\begin{math}e_5\end{math}\} in frame \{\begin{math}frame\end{math}\ \begin{math}number\end{math}\}." This change hopes to guide VideoLLaMA2 in determining where each event is occurring, allowing it to better learn with this added information. 
    \item \textbf{Providing event count.} Given poor performances in previous variations, we want to see if directly telling the model how many events to predict will improve its accuracy. While such a number will not be provided in new data, it helps identify current problems with the model. An increase in perform will highlight the initial model's inability to predict, by itself, the number of events in a video. To do so, we append a clause to the start of each prompt, resulting in the format: ``Given that there are \{\begin{math}x\end{math}\} tennis actions in this video, what is happening in the tennis video?"
    \item \textbf{Providing self-predicted event count.} Given the drastic increase in performance in Variation 6, we attempt to combine Variation 2 with Variation 6, allowing the model to make predictions on new data without any external help. We first make predictions on the number of events as per Variation 2, and feed these numbers into our model in Variation 6. 
    
\end{enumerate}

\subsection{Edit score} 

In this task, we will be using normalised segmental edit score \cite{lea2016learning} as our metrics to assess the performance of our models. The edit score is calculated using the normalised Levenshtein Distance between the prediction and the true value, which is reflected in the formula below. In essence, it measures the number of additions, deletions or replacements needed to change the predicted string into the true string. In this case, we treat each individual event as a single word. 

\[ \text{Edit Score} = (1 - \frac{\text{Event additions, deletions or replacements needed}}{\text{Number of events in longer label}}) * 100 \]\\

\par Tables 5 and 6 consolidate the performances of all the variations thus far. All experiments were fine-tuned to 10 epochs unless otherwise specified, as it was found that this was an optimal balance in determining the performance of the model while not spending too much time and resources.

\begin{table}[h!]
\begin{center}
\begin{tabular}{ |c|c| } 
 \hline
 \textbf{Variation} & \textbf{Edit score} \\
 \hline
 Default (8 epochs) & 30.2 \\
 Default (10 epochs) & 34.4 \\
 Default (20 epochs) & 33.1 \\
 Default (50 epochs) & 28.0 \\
 Audio & 25.3 \\
 Increased frame sampling & 39.7 \\
 Providing frame numbers & 33.9 \\
 Providing event count & 49.8 \\
 Providing self-predicted event count & 30.0 \\
 
 Benchmark (on higher granularity task) & 82.1 \\

 \hline
\end{tabular}
\caption{Results for different variations}
\end{center}
\end{table}

\begin{table}[h!]
\begin{center}
\begin{tabular}{ |c|c| } 
\hline
 Accuracy & 0.36 \\
 \hline
 Average difference in number of events & 1.31 \\
 \hline
 Average correct number of event & 3.90 \\
 \hline
 Standard Deviation of correct number of events & 3.46 \\
\hline
\end{tabular}
\caption{Results for Event Counting}
\end{center}
\end{table}

\subsection{STC Connector Isolation}
Due to the importance of maintaining spatial temporal information in the task, we wanted to assess the effectiveness of VideoLLaMA2's STC connector in doing so. In our preliminary experiment, we mimicked the VideoLLaMA2 model up till the STC connector. The frames are passed into the CLIP vision encoder, which is frozen, before being passed to the STC connector. Similarly, the outputs are flattened. However, instead of passing it to the rest of the VideoLLaMA2 model, we simply added a single hidden layer and an output layer. It is worth noting that the STC connector layers here are retrained from scratch. This experiment was done on the single event classification to allow for a simple output layer and aims to identify if the features being extracted by the STC connector alone are sufficient in determining the event in each video. 

\begin{figure}[h]
    \begin{center}
    \includegraphics[width=0.5\linewidth]{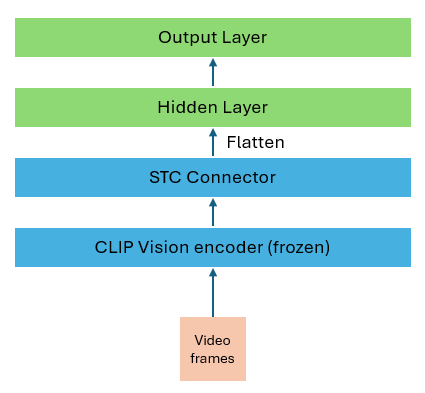}
    \caption{Simplified STC Connector model}
    \label{fig:enter-label}
    \end{center}
\end{figure}

\par Similar to previous experiments, we trained this model on 10 epochs. In the end, this simplified STC connector model performed poorly, attaining an accuracy of only 0.018, which was comparable to randomly guessing the class. This posed the question if the current STC connector was truly capable of capturing the spatial temporal information it was supposed to. Without another connector to compare to, we cannot conclude this poor performance is the fault of the STC connector, but this result suggests that improvements can be made to better capture the spatial temporal information needed for predictions.

\subsection{Frame sampling}
Unlike traditional models, many open-source MLLMs use frame sampling. In the case of VideoLLaMA2, the model samples only 8 frames, and has been trained sampling only 8 frames. While this might work in tasks requiring only a general understanding of videos, this might cause a lot of important information to be lost. \cite{fu2024video} When the number of frames sampled increased to 32, the edit score of the model noticeably increased as well, reaffirming that such limited sampling methods could pose a hindrance to video sequence identification.  A better approach for MLLMs would be to sample every fixed number of frames to ensure consistent information density \cite{ren2023testa}, instead of only sampling a fixed number of frames regardless of the length of the video. To test this hypothesis, we modified the MLLM to take in all the frames in the video and carried out the same sequence identification task with the same default prompt. 

Surprisingly, the edit score failed to improve, scoring only \textbf{28.6}, similar to our over-fitted model when trained on 50 epochs. This result confirms that frame sampling is not the main contributing factor to the model's poor performance, and is likely due to something within the model itself.
Noting how this change did not impact the edit score significantly, we continued working with our frame sampling method of 32 frames which gave the highest edit score so far.

\subsection{Using other models with VideoLLaMA2}

In this chapter, through collaboration with research assistant, Liu Zhaoyu, we analyse how the use of other models can be used to enhance the performance of our MLLM. This includes the use of models such as ball tracking, player detection and pose detection.

\subsubsection{Player bounding boxes}
\par First, we used an open source object detection model to detect the people in each video, getting each of their bounding boxes. However, since this gave rise to many detections of referees and spectators, we had to add an extra algorithm to extract the players only. To do so, we used an open source court detection algorithm to detect the different keypoints of the tennis court in the video. We then select the closest people detections from each end of the court and extract those as our bounding boxes for both players. If a suitable bounding box is not found for either player, the coordinates will all default to (-1, -1).

\par The photos below show some examples of our results.

\begin{figure}[H]
    \begin{center}
    \includegraphics[width=0.8\linewidth]{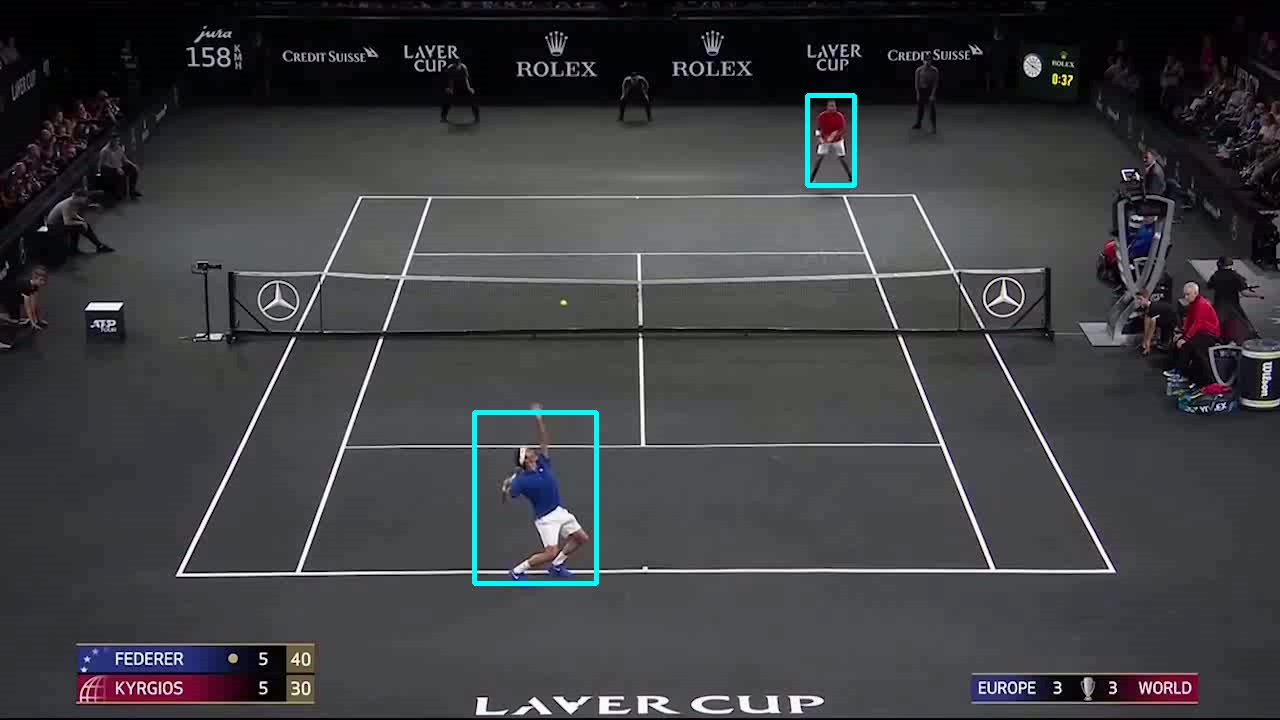}
    \caption{Sample player bounding boxes 1}
    \label{fig:enter-label}
    \end{center}
\end{figure}

\begin{figure}[H]
    \begin{center}
    \includegraphics[width=0.8\linewidth]{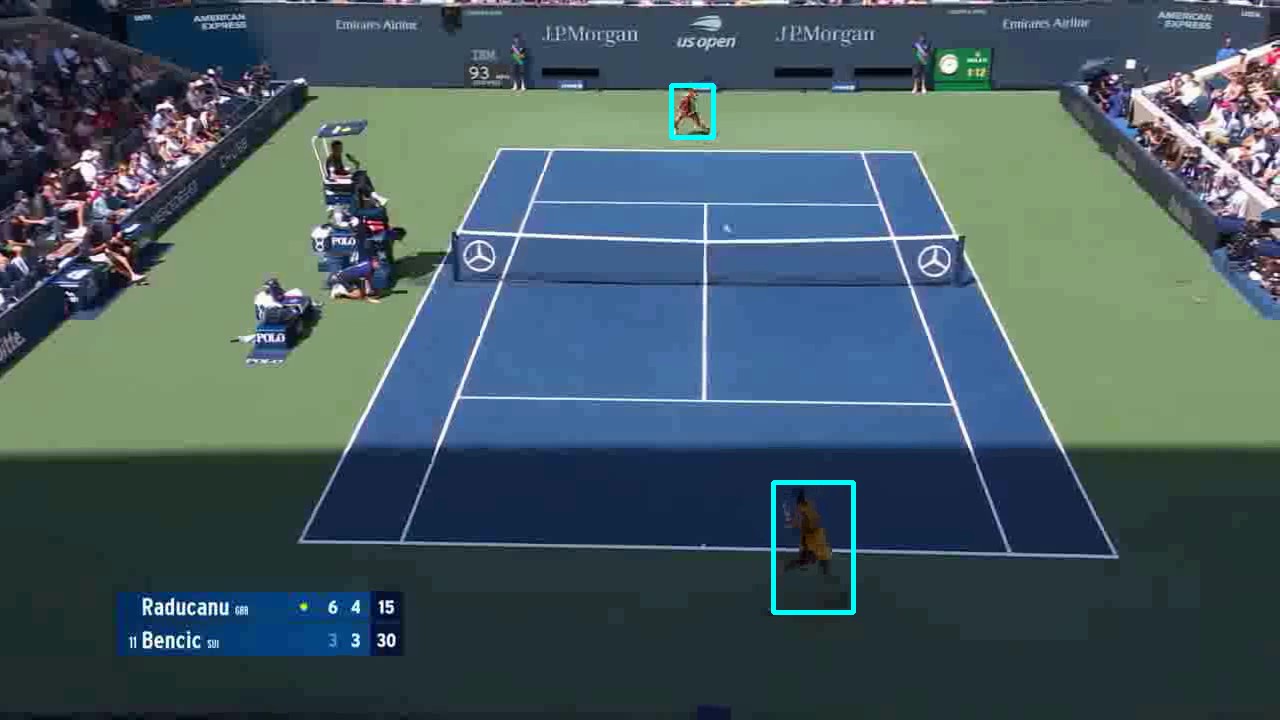}
    \caption{Sample player bounding boxes 2}
    \label{fig:enter-label}
    \end{center}
\end{figure}

\par After extracting these player bounding boxes, we tried three different methods to incorporate this new information:

\begin{enumerate}
\item \textbf{Including bounding boxes in video.} We created a new training and testing dataset by drawing the bounding boxes of all players in the videos themselves, as shown in the photos above. This new dataset is then used for fine-tuning on the default VideoLLaMA2 under the same prompts and hyperparameters (32 frames) above. The idea was to provide more information in the video for the MLLM, in the hope that this would help it to extract features better. In the end, this gave rise to a sharp drop in edit score from the MLLM, falling from the initial 39.7 down to \textbf{18.2}, more than a 50\% drop.

\item \textbf{Including bounding boxes in prompt.} Instead of giving this new information in the video, we tried providing it in the prompt. To do so, we came up with a new prompt structure as follows: \textit{Given this list of the far player's bounding box in each frame in format (\begin{math}x_1, y_1, x_2, y_2 \end{math}): \{far\_bbox\}, and this list of the near player's bounding boxes in each frame in format (\begin{math}x_1, y_1, x_2, y_2 \end{math}): \{near\_bbox\}, describe all the tennis actions in the video.} 

In the prompt, \begin{math}x_1, y_1\end{math} corresponds to the bottom left corner of the bounding box and \begin{math}x_2, y_2\end{math} corresponds the top right corner of the bounding box.

This experiment showed great promise, resulting in a sharp increase in the edit score of the model, bringing it up to \textbf{61.3}, the highest edit score we have achieved thus far. 

\item \textbf{Including bounding boxes in both video and prompt.} Lastly, we attempt to combine both the first and second approach, using the prompt in point 2 together with videos containing the bounding boxes of the players. 

Ultimately, our model's edit score fell slightly from 61.3 to \textbf{59.1}. 

\end{enumerate}

\par In general, our results indicate that drawing bounding boxes in the video itself had an overall detrimental effect on the model's learning. We posit that this is likely due to the new lines disrupting the vision encoder, which has been primarily trained on videos without them. These lines also temporarily cover the ball in certain frames, and overlap with the tennis courts, which might cause even more confusion for the model. As such, we leaned away from editing videos directly moving forward, noting this poor performance. 

\subsubsection{Adding ball and court information}
\par As mentioned in the previous section, we had made use of an open source court detection model to determine the far and near players. After seeing the success we had in providing the player bounding box coordinates, we decided that providing the information of the court coordinates might give extra context and aid the model in performing even better. The court detection model gave 4 coordinates - the 4 corners of the tennis court, which we then fed into our MLLM prompt. 

The images below show some of the results of our court detection model, with the court's four corners denoted by red X's.

\begin{figure}[H]
    \begin{center}
    \includegraphics[width=0.8\linewidth]{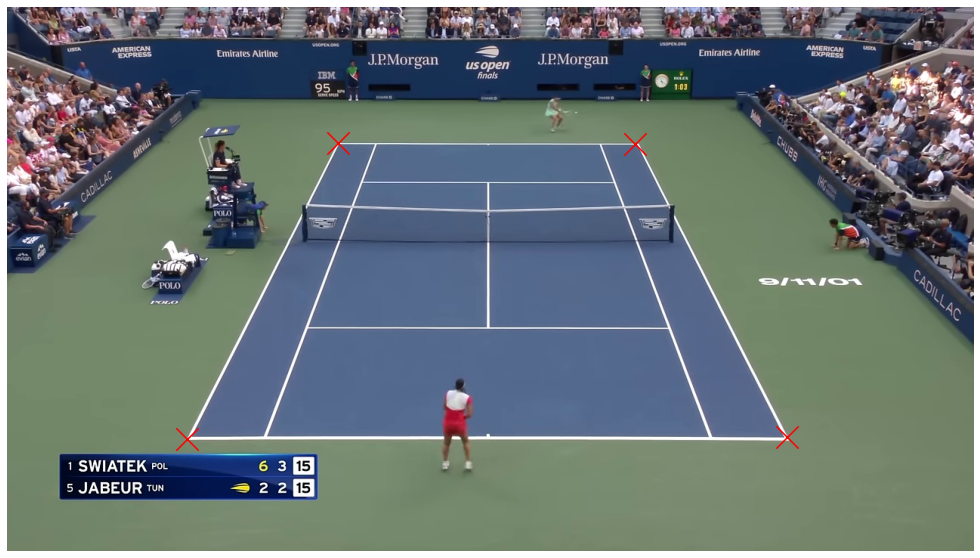}
    \caption{Sample court detection result 1}
    \label{fig:enter-label}
    \end{center}
\end{figure}

\begin{figure}[H]
    \begin{center}
    \includegraphics[width=0.8\linewidth]{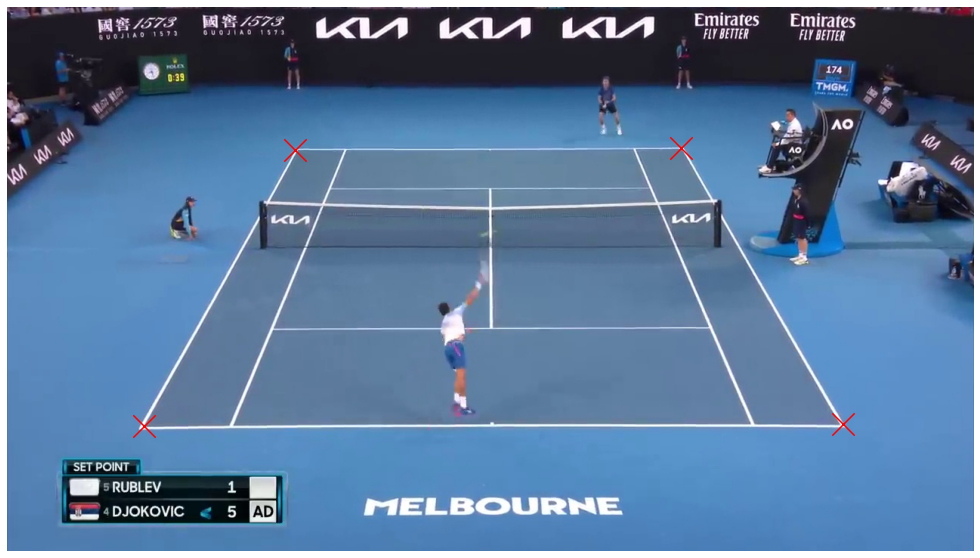}
    \caption{Sample court detection result 2}
    \label{fig:enter-label}
    \end{center}
\end{figure}

We took it one step further and also found a ball detection model for tennis. In a similar fashion, we used the open source model to detect the ball in each frame of the video. Since the tennis balls are small in each frame, we only took 1 set of (x, y) coordinates, denoting the center of the ball. 

Once again, we visualise these results in the pictures below, which also show some promising results. 

\begin{figure}[H]
    \begin{center}
    \includegraphics[width=0.8\linewidth]{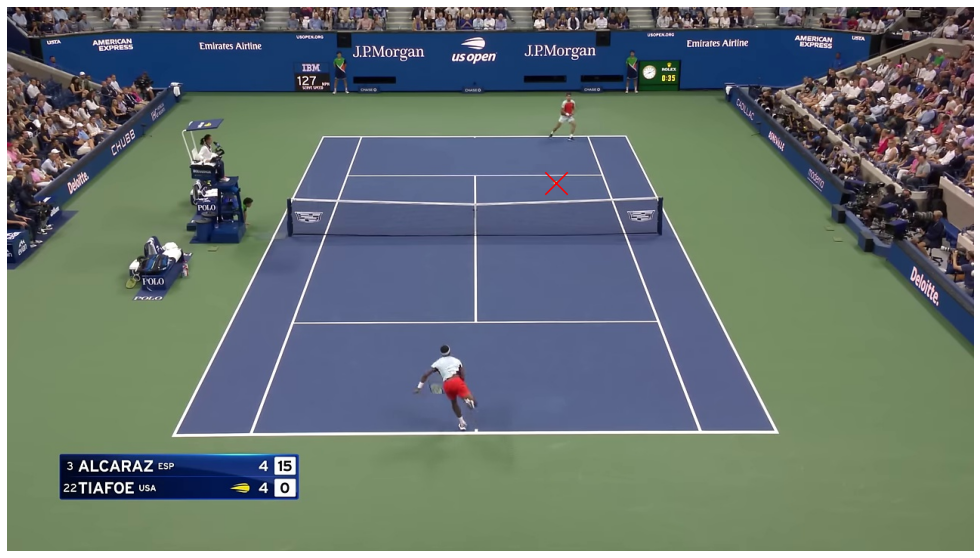}
    \caption{Sample ball detection result 1}
    \label{fig:enter-label}
    \end{center}
\end{figure}

\begin{figure}[H]
    \begin{center}
    \includegraphics[width=0.8\linewidth]{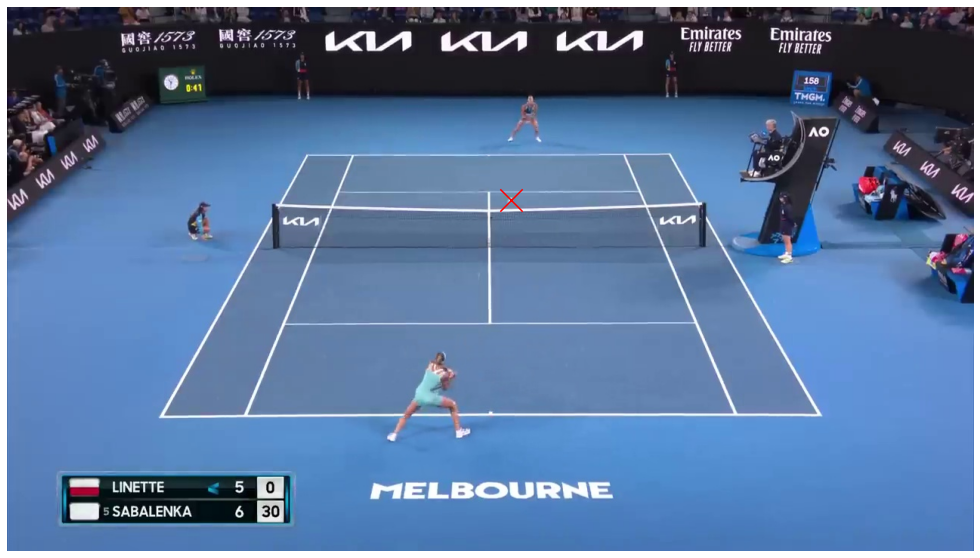}
    \caption{Sample ball detection result 2}
    \label{fig:enter-label}
    \end{center}
\end{figure}

Same as the player bounding boxes, if the court or ball is not detected, all coordinates will be defaulted to -1 for the model differentiate. 

To pass this information to the model, we again modify our prompt to the following: \textit{Given this list of the far player's bounding box in
format (\begin{math}x_1, y_1, x_2, y_2\end{math}): \{far\_bbox\}, and this list of the near player's bounding boxes in format (\begin{math}x_1 \end{math}, \begin{math}y_1 \end{math}, \begin{math}x_2 \end{math}, \begin{math}y_2 \end{math}): \{near\_bbox\}, and this list of the tennis ball's coordinates in format (x, y): \{ball\}, and given the following court dimensions in the form of (\begin{math}x_1 \end{math}, \begin{math}y_1 \end{math}, \begin{math}x_2 \end{math}, \begin{math}y_2 \end{math}, \begin{math}x_3 \end{math}, \begin{math}y_3 \end{math}, \begin{math}x_4 \end{math}, \begin{math}y_4 \end{math}): \{court\}, describe the tennis actions in the video.}

However, when passing this prompt into our model, we faced the issue of not having enough RAM in our computing cluster to fine-tune the MLLM. Due to the large number of frames some of the longer rallies have, the number of coordinates that are passed on become extremely huge as well. In order to solve this, we halved the number of coordinates passed by passing only information from every other frame. Using this approach and the same setup otherwise as in the above section, the edit score continued to increase from 61.3 before to \textbf{76.0}.

This result further proves our hypothesis that providing information from other models into the prompt of the MLLM can greatly boost its accuracy, urging us to attempt other models, especially those that might provide more information.

\subsubsection{Pose estimation}

For our last experiment, we decided to make use of a pose estimation model, giving more comprehensive information on the player model than just their bounding boxes. Our pose estimation model gives us 17 different keypoints on each person, following the format of the COCO-Pose Dataset. 

Due to the limitation of our RAM and increased number of coordinates, we had to increase the interval between frames, providing information in fewer frames instead to fit our computing constraints. 

Some example results of the pose estimation model are shown below:

\begin{figure}[H]
    \begin{center}
    \includegraphics[width=0.8\linewidth]{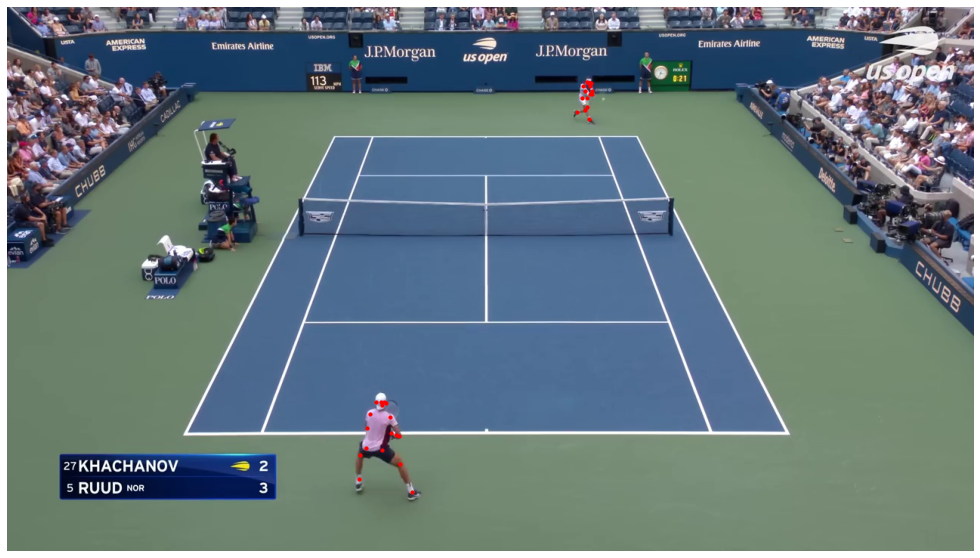}
    \caption{Sample pose detection result 1}
    \label{fig:enter-label}
    \end{center}
\end{figure}

\begin{figure}[H]
    \begin{center}
    \includegraphics[width=0.8\linewidth]{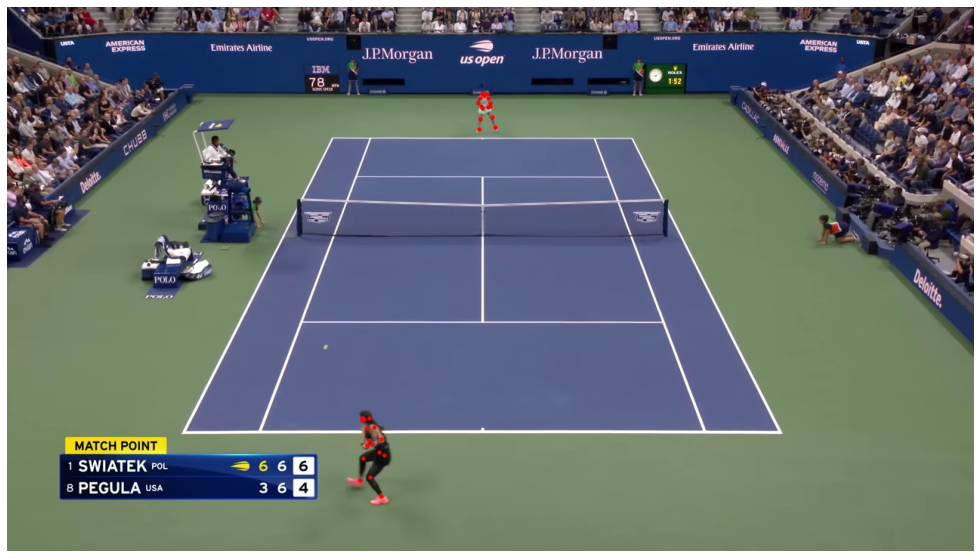}
    \caption{Sample pose detection result 2}
    \label{fig:enter-label}
    \end{center}
\end{figure}

Our initial visualisations show that our pose estimation model also does have decent results.

First, we tried feeding all the keypoints into the model. With some basic searching, we found that the minimum interval is 20 frames. We then replaced the bounding box line in our prompt with the following line: \textit{Given this list of the far player's keypoints: \{far\_keypoints\}, and this list of the near player's keypoints: \{near\_keypoints\}}. This experiment resulted in a \textbf{50.7} edit score, much lower than our initial edit score. 

We hypothesise that this might be due to the larger interval, not allowing the model to identify the actions properly. To test this, we decided to conduct another experiment, only providing 4 keypoints - each players' left and right hands, and their left and right feet. This allowed us to decrease the interval to every 5 frames instead. Using the same prompt and same training, this brought the edit score to \textbf{57.6}, reaffirming our hypothesis. Nonetheless, this still caused a significant decrease from our original edit score of 76.0, implying that this additional information did not improve our model at the cost of a longer interval. The table below summarises all the results in this section.

\begin{table}[h!]
\begin{center}
\setlength\extrarowheight{3.5pt}
\begin{tabular}{ |c|c| } 
 \hline
 \textbf{Model} & \textbf{Edit score} \\
 \hline
 Default (Benchmark) & 39.7 \\
 Player bounding boxes in video & 18.2 \\
 Player bounding boxes in prompt & 61.3 \\
 Player bounding boxes in both video and prompt & 59.1 \\
 Player bounding boxes, court and ball detection (every 2 frames) in prompt & 76.0 \\
 \makecell{Player's 17 keypoints, court and ball detection (every 20 frames)\\ in both video and prompt} & 50.7 \\
 \makecell{Player's 4 keypoints, court and ball detection (every 5 frames)\\ in both video and prompt} & 57.6 \\
 \hline
\end{tabular}
\caption{Results for varying methods of merging different models}
\end{center}
\end{table}

\subsection{Vision Encoder}
Upon further reviewing the architecture of VideoLLaMA2, we began to question the freezing of the vision encoder. The current vision encoder might not be trained to pick up features specific to tennis, given that the original CLIP model was trained on a wide range of rather general items \cite{radford2021learning}. Thus, the features passed downstream might not be as useful as it should be. To fix this, we decided to experiment fine-tuning the VideoLLaMA2 model while unfreezing the vision encoder layer, allowing the vision encoder layer to learn from the tennis data on what features are useful. We first tested this on single event classification, a simpler task, to assess its initial effectiveness. 
\subsubsection{Unfrozen visual encoder} 
Due to the increased number of parameters and the time limit set on the shared computing cluster, we decreased the number of epochs to 4. However, our results showed that the accuracy of the model fell drastically, from the initial 0.41 to 0.075. To investigate further, we noticed that during training, the loss of this model fell much quicker as compared to the initial model. For comparison, the loss of the new model fell to below 0.1 in the first epoch already, while the initial model's loss was still above 0.1 even after 10 epochs. Such low losses during training but poor performances on the test set implies some form of overfitting. The unfrozen encoder layer provided the model with more flexibility, which could have caused it to be more prone to overfitting \cite{hawkins2004problem}. 
\par We tried fine-tuning the new model again but this time with only 1 epoch, hoping to prevent the overfitting. However, the model continued to perform poorly, achieving a slightly better peformance of 0.087. With the poor results and a lack of papers to provide more information on this direction, we were forced to conclude that fine-tuning the model as a whole, including the vision encoders, was not the way forward. 
\subsubsection{CLIP Model} 
Intuitively, fine-tuning the vision encoder should allow the model to better learn what useful features to extract from the videos, leading to better performance. To test this theory, we fine-tuned the CLIP model itself on our dataset to gauge its performance. In this experiment, we used the full CLIP model, inclusive of text embedding layers, to make use of the in-built output layer which provides the logits of each class that we can use for predictions and loss calculations. After all, doing so should still fine-tune the vision encoder layers in extracting useful features from the videos. Again, due to the large number of trainable parameters in the full CLIP model, we first experimented fine-tuning on 1 epoch, which produced a promising accuracy of 0.43 on the test data, outperforming the VideoLLaMA2 model already. With this result, fine-tuning the vision encoder did seem like a viable strategy to employ again. 
\par Despite the sparse number of papers exploring fine-tuning MLLMs due to their recency, we found a recent paper that shared an improved performance when fine-tuning the CLIP model separately and reintegrating it into the MLLM \cite{panos2024imperfect}, leading us to go down this direction. 
\subsubsection{CLIP fine-tuned separately} 
We loaded the new CLIP vision encoder layers, fine-tuned on 1 epoch, into VideoLLaMA2 before freezing the layer as before, and fine-tuned the rest of VideoLLaMA2. We first experimented with 10 epochs, and the results exceeded the initial model's significantly, from 0.41 to 0.55. We, again, noticed that the training loss of this model fell quickly and approached 0 nearing the 8th epoch. As such, we tried the experiment again with 6 epochs and found that the model's accuracy increased slightly to 0.56, implying that training past 10 epochs might no longer be helpful. Nonetheless, this step up in accuracy indicated a big step in the correct direction.
\par The results mentioned above are consolidated in the table below. 

\begin{table}[h!]
\begin{center}
\begin{tabular}{ |c|c| } 
 \hline
 \textbf{Model} & \textbf{Accuracy} \\
 \hline
 Default & 0.41 \\
 Unfrozen visual encoder (1 epoch) & 0.087 \\
 Unfrozen visual encoder (4 epoch) & 0.075 \\
 CLIP Model (1 epoch) & 0.43 \\
 CLIP fine-tuned separately (6 epoch) & 0.56 \\
 CLIP fine-tuned separately (10 epoch) & 0.55 \\

 \hline
\end{tabular}
\caption{Results for different variations of CLIP fine-tuning}
\end{center}
\end{table}

\subsubsection{Sequence Identification} After verifying that fine-tuning the CLIP vision model separately has significant improvements to the model on single event classification, we then use the same method for our sequence identification task. The same fine-tuned CLIP model is loaded into VideoLLaMA2 and frozen before fine-tuning the other layers using LoRA on the sequence identification task. We used 32 frames as a comparison in our experiment. As expected, we achieved a significant jump in edit score, from the initial 39.7 to \textbf{54.6}. 

\newpage
\section{Analysis of initial investigations}

Based on our current results, below are some possible conclusions we can draw.

\subsection{Textual understanding vs video understanding}

When asked to identify the sequence of events with just the video, VideoLLaMA2 performs poorly, achieving about half the edit score of our benchmark, regardless of any modification we make to the model. However, when we pass information from the video in textual format, all information contained solely in the video, the model begins to perform significantly better. This might be due to the model's ability to find patterns in textual information passed to it as compared to other modes such as videos. In comparison, when all frames of the video are passed to the model, the edit score is significantly lower, further supporting the point that the model's vision backbone has a lot to improve on.

We also identify that VideoLLaMA2 performed well in recognising patterns in the rally. All rallies correctly began with a serve for \begin{math}
    e_3
\end{math} and ended with a last for \begin{math}
    e_5
\end{math}. All rallies also correctly alternated between the near and far player. This is all information that was probably obtained through textual understanding, given the perfect accuracy rate. 

During our literature review, we also found research supporting this claim, reporting that MLLMs did well in textual reasoning but performs much more poorly in scenario-based reasoning and matching frames to sporting rules.

In general, this points to a need to change the current vision backbone of VideoLLaMA2 in order to achieve better results on this sequence identification task.

\subsection{Event counting as benchmark}
VideoLLaMA2 performs poorly on determining the number of events in a video without guidance. The jump in edit score when given the number of events in the video highlights this flaw in the current iteration of VideoLLaMA2. As mentioned in our literature review, many benchmarks mainly assess MLLMs' abilities to get a general understanding of videos. The ability to identify the number and type of different events in a video is an important improvement we can add to current benchmarks. Such tasks are significantly more complex, requiring more research to improve on. Yet, such a benchmark is highly relevant to real life and could present significant opportunities in the future. 

\subsection{Separate vision encoder fine-tuning}
Our results suggest an underexplored approach in fine-tuning MLLMs, separately fine-tuning the vision encoder before freezing it again in the model for another round of fine-tuning. While only experimented on one model, this process showed great promise in our initial sequence identification task, improving our edit score from 39.7 to 54.6. Future fine-tuning processes can consider such an approach to improve their accuracy as well. 

\subsection{Comparison with traditional models}
While the results for merging MLLMs with other more basic traditional models shows promise in more complex tasks, the current results for both single shot and rally predictions indicate that VideoLLaMA2, especially on its own, cannot compare to the current benchmark scores we have. This indicates that there is more to be done and improved on in the field of MLLMs to allow them to function on their own and match up to other traditional models.
\newpage
\section{Conclusion}
\subsection{Limitations and future work}
Next, we discuss some possible reasons for the poor performance of VideoLLaMA2 in our experiments, and some ideas we have for future work.

\subsubsection{Too much or too little information}
From our results, there appears to be a sweet spot as to how much information we should give a model. Increasing frame sampling from 8 to 32 increases the edit score but going overboard and sampling every single frame decreases the model's performance. We see a similar trend when combining with other models, where giving bounding box information every 2 frames quite significantly outperforms keypoint information every 5 frames. While frame interval might play a role, the huge jump in edit score might suggest providing bounding box information is more useful than keypoints. This indicates that with some information, the model does well, but too much or too detailed information causes its performance to fall. 

The possibility of the existence of a sweet spot and finding this balance between too much and too little information is an interesting aspect of our model and could be useful to look into given more time. 

\subsubsection{STC Connector}
Our result with the STC connector alone poses the possibility that the current STC connector model used in VideoLLaMA2 is not fit for our task of sequence identification, performing poorly just on tennis action identification alone. While this might not be the reason for its poor performance, it is still a consideration to be had.  

With more time, we would like to experiment with other possible variations that can be used as an STC connector, though this would be more complex.

\subsubsection{Vision encoder}
The vision encoder is completely frozen, and thus not changed during the fine-tuning process. The overly-generalised vision encoder could possibly not be capable of extracting the correct features in the tennis videos. Our experiment fine-tuning the vision encoder shows its capabilities to outperform the MLLM, indicating that such a step could aid the MLLM in making better predictions. The initial experiment to integrate this separately fine-tuned CLIP encoder back into VideoLLaMA2 also presented promising results, providing a possible direction to explore in improving the fine-tuning process for MLLMs. Interestingly, our experiments also indicate fine-tuning MLLMs fully, unfreezing vision encoders, results in significantly poorer performances, while fine-tuning the two separately significantly boosts their combined performance.

Given the relative novelty of this approach, more research and experimentation is needed to support our findings. Future work could include fine-tuning different MLLMs on different tasks using this approach and determining its impact on the model's accuracy.

\subsubsection{LoRA Fine-tuning}
LoRA was adopted due to our limitations in hardware and its low VRAM consumption. However, this decrease in trainable parameters can fall short of the effectiveness of full fine-tuning. \cite{hu2021lora} A full fine-tuning on all trainable parameters could present a higher edit score than the one obtained. 

With more resources, we propose carrying out these experiments with the full VideoLLaMA2 model to ascertain its performance on sequence identification. This also includes the other experiments such as giving more frequent information in the prompt instead of doing frame intervals in our experiments, to see the maximum potential of the model.

\subsubsection{Other models}
We also adopted VideoLLaMA2 due to its lightweight LoRA fine-tuning. Given more time and resources, more experiments could have been carried out on different models to see if they share the same problems and tendencies. Doing so could further support many of our findings, such as accuracy improvements due to separate fine-tunings or MLLM's superior textual reasoning compared to its video reasoning. This will help us better generalise to MLLMs as a whole and identify general weaknesses and strengths for current MLLMs. 

\subsection{Summary}
All in all, our research shows that there is a still a lot of improvement to be had for MLLMs, especially in the area of sequence of identification. We also posit a better way for fine-tuning MLLMs through two separate fine-tuning processes to maximise the model's accuracy. Our research also indicates that MLLM's have powerful textual processing, making it useful for combining with outputs from other models, but is usually unable to extract such information on its own from its vision backbone. Nonetheless, MLLMs present exciting opportunities, especially in the world of sports, and are making steady progress to do so effectively.

\newpage
\pagenumbering{roman}
\setcounter{page}{4}
\addcontentsline{toc}{section}{\protect\numberline{}References}
\bibliographystyle{apacite}
\renewcommand{\arraystretch}{1.5}
\bibliography{references} 

\newpage

\addcontentsline{toc}{section}{\protect\numberline{}Appendix A: Github repository}
\noindent
\Large{\textbf{Appendix A: Github repository}}
\\
\large{All codes useful for future work have been re-formatted and stored in this Github repository: \\
\url{https://github.com/bigcrushes/videollama2_tennis}}

\end{document}